\documentclass[conference]{IEEEtran}
\IEEEoverridecommandlockouts

\usepackage{cite}
\usepackage{amsmath,amssymb,amsfonts}
\usepackage{algorithmic}
\usepackage{graphicx}
\usepackage{textcomp}
\usepackage{xcolor}
\def\BibTeX{{\rm B\kern-.05em{\sc i\kern-.025em b}\kern-.08em
    T\kern-.1667em\lower.7ex\hbox{E}\kern-.125emX}}

\makeatletter
\newcommand{\linebreakand}{%
  \end{@IEEEauthorhalign}
  \hfill\mbox{}\par
  \mbox{}\hfill\begin{@IEEEauthorhalign}
}
\makeatother

\usepackage{xcolor}
\usepackage{hyperref} 
\usepackage{amsmath,amssymb}
\usepackage{bm}
\usepackage{booktabs}
\usepackage{comment}
\usepackage{multirow}
\usepackage{framed}

\usepackage{url}
\makeatletter
\newcommand{\figcaption}[1]{\def\@captype{figure}\caption{#1}}
\newcommand{\tblcaption}[1]{\def\@captype{table}\caption{#1}}
\makeatother

\newcount\K
\def\bla#1{
\K=0 \loop\ifnum\K<#1
{\textcolor[gray]{0.9}{{\it bla bla bla bla bla bla bla bla bla bla bla bla bla bla bla}}}
\advance\K by1\repeat
}

\newcommand{\todo}[1]{
\ifx#10
\textcolor{red}{$0.00_{\pm 0.00}$}
\else
\textcolor{red}{#1}
\fi
}

\begin{document}

\title{Multi-Point Positional Insertion Tuning\\
for Small Object Detection}

\author{\IEEEauthorblockN{Kanoko Goto}
\IEEEauthorblockA{\textit{Dept. of Computer Science} \\
\textit{Institute of Science Tokyo}\\
Tokyo, Japan.}
\and
\IEEEauthorblockN{Takumi Karasawa}
\IEEEauthorblockA{\textit{Dept. of Systems and Control Engineering} \\
\textit{Institute of Science Tokyo}\\
Tokyo, Japan.}
\and
\IEEEauthorblockN{Takumi Hirose}
\IEEEauthorblockA{\textit{Dept. of Computer Science} \\
\textit{Institute of Science Tokyo}\\
Tokyo, Japan.}
\linebreakand
\IEEEauthorblockN{\hspace{15pt}Rei Kawakami}
\IEEEauthorblockA{\textit{\hspace{15pt}Dept. of Systems and Control Engineering} \\
\textit{\hspace{15pt}Institute of Science Tokyo}\\
\hspace{15pt}Tokyo, Japan.}
\and
\IEEEauthorblockN{Nakamasa Inoue\hspace{43pt}}
\IEEEauthorblockA{\textit{Dept. of Computer Science\hspace{43pt}} \\
\textit{Institute of Science Tokyo\hspace{43pt}}\\
Tokyo, Japan.\hspace{43pt}}
}
\maketitle
\begin{abstract}
Small object detection aims to localize and classify small objects within images. With recent advances in large-scale vision-language pretraining, finetuning pretrained object detection models has emerged as a promising approach.
However, finetuning large models is computationally and memory expensive.
To address this issue,
this paper introduces multi-point positional insertion (MPI) tuning, a parameter-efficient finetuning (PEFT) method for small object detection.
Specifically, MPI incorporates multiple positional embeddings into a frozen pretrained model, enabling the efficient detection of small objects by providing precise positional information to latent features.
Through experiments, we demonstrated the effectiveness of the proposed method on the SODA-D dataset. MPI performed comparably to conventional PEFT methods, including CoOp and VPT, while significantly reducing the number of parameters that need to be tuned.
\end{abstract}


\section{Introduction}
Object detection has become a crucial component in real-world applications such as autonomous driving and surveillance owing to the remarkable advancements in deep neural networks.
To accurately detect small objects within images, various methods have been developed, such as super-resolution methods~\cite{noh2019bffb,bai2018sod-mtgan}, similarity learning~\cite{yuan2023cfinet,kim2021lprmemory,wu2020selfmimic}, and context exploration~\cite{Zhang2023localglobal}.
However, detecting extremely small objects is still challenging, mainly because of the insufficiency of training data, as manual annotation of the bounding boxes for these objects is time-consuming and costly.

To address this issue, a recent trend has relied on large-scale pretraining. Specifically, for object detection, some studies have proposed vision-language models pretrained on large-scale datasets, such as GLIP~\cite{li2021glip, zhang2022glipv2} and Grounding DINO~\cite{liu2023groundingdino, zhao2024open}.
These models are open-set object detectors capable of accepting natural language text or a sequence of object names as inputs.
They can also be adapted as closed-set object detectors for detecting objects within a predefined category set by finetuning them on limited labeled datasets~\cite{liu2023groundingdino, zhao2024open}.
Therefore, they are expected to be effective for small object detection.

\begin{figure}
\centering
\includegraphics[width=\linewidth]{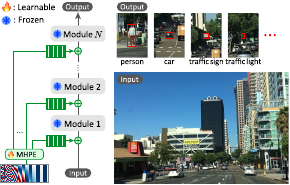}
\caption{
Multi-point positional insertion (MPI) tuning for small object detection.
MPI tuning inserts positional embeddings at multiple points in a frozen pretrained model through a learnable multi-head positional encoder (MHPE). This figure illustrates a frozen object detection model with $N$ sequential modules for simplicity.
}
\label{fig:top}
\end{figure}

When finetuning large models, a primary challenge remains in terms of parameter efficiency, as optimizing a large number of parameters is computationally and memory expensive.
To improve parameter efficiency,
adapter tuning~\cite{long2024multiway, zhou2024automatic, zhang2024test, gao2024adapter, nlp-adapter} and
prompt tuning~\cite{zhou2022coop, zhou2022cocoop, jia2022vpt, Lin2024Visual, Xu2024Enhanced, wang2024cophtc} are known to be effective.
These methods insert lightweight learnable modules into a frozen pretrained model, allowing the model to adapt to new tasks with a minimal increase in the number of learnable parameters while avoiding overfitting.

Inspired by these studies, this paper introduces a novel parameter-efficient finetuning (PEFT) method for small object detection.
Specifically, we propose multi-point positional insertion (MPI) tuning, which incorporates multiple positional embeddings into a pretrained frozen model, as shown in Figure~\ref{fig:top}, enabling the efficient detection of small objects by providing precise positional information to latent features.
In our experiments, we demonstrated the effectiveness and parameter efficiency of MPI tuning on the SODA-D dataset~\cite{cheng2023soda}.
We observed that MPI tuning performs comparably to conventional PEFT methods, while reducing the number of learnable parameters.

\section{Related work}

\noindent \textbf{Object detection.}
Over the last decade, numerous object detection models have been proposed~\cite{lin2017retinanet, tian2019fcos, ren2015fasterrcnn, sun2021sparsercnn, carion2020detr}.
There have been two major architectures: convolutional architectures, {\it e.g.}, RetinaNet~\cite{lin2017retinanet} and Sparse RCNN~\cite{sun2021sparsercnn}, and transformer-based architectures, {\it e.g.}, DETR~\cite{carion2020detr} and Deformable DETR~\cite{zhu2020deformabledetr}.
Recently, vision-language pretrained models such as GLIP~\cite{li2021glip, zhang2022glipv2} and Grounding DINO (GDINO)~\cite{liu2023groundingdino, zhao2024open} have demonstrated effectiveness in open-set object detection and visual grounding.
For small object detection, convolutional architectures, such as CFINet~\cite{yuan2023cfinet} using coarse-to-fine region proposals, remain the primary approach~\cite{jinyu24icassp, Li2024SOD, Zhu2023Small, Zhang2023Dynamic}.

\noindent \textbf{PEFT.}
Adapter tuning inserts lightweight learnable modules into a frozen pretrained model~\cite{long2024multiway, zhou2024automatic, zhang2024test, gao2024adapter, nlp-adapter, Otake2023Parameter}.
For instance, encoder adapter tuning~\cite{nlp-adapter} incorporates small multilayer perceptrons (MLPs) into each encoder layers.
Layer adapter tuning~\cite{Otake2023Parameter} inserts small modules between each layer and the downstream head.
Prompt-based finetuning has also garnered attention because of its success in the field of natural language processing~\cite{zhou2022coop, zhou2022cocoop, jia2022vpt, Lin2024Visual, Xu2024Enhanced, wang2024cophtc}.
Examples include context optimization (CoOp)~\cite{zhou2022coop} for text prompt tuning and visual prompt tuning (VPT)~\cite{jia2022vpt}.
This study focuses on PEFT for small object detection, where encoding precise spatial positions within images is crucial.

\section{Method}

This section presents MPI tuning, a PEFT method for small object detection.
MPI tuning inserts positional embeddings into multiple points within a frozen pretrained model.
This approach provides precise positional information for latent features, enabling efficient adaptation for detecting small objects.

\subsection{Notation and settings}

Object detection aims to localize and classify objects within images.
Specifically, the objective is to provide bounding boxes and categories for each object given an input image and predefined object categories.
This study discusses the parameter efficiency of finetuning given a pretrained object detector $f$.
We assume that $f$ is a deep neural network and involves latent features.
Given an input $\bm{x}$, we denote by $\mathcal{H}(\bm{x}) = \{\bm{h}_{i}(\bm{x})\}_{i=1}^{N}$ the set of latent features in the neural network, where $N$ is the number of latent features.

\begin{figure}
\centering
\includegraphics[width=\linewidth]{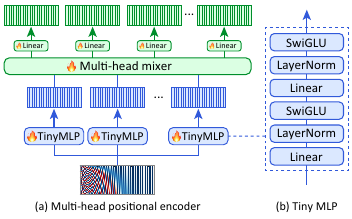}
\vspace{-12pt}
\caption{(a) Multi-head positional encoder consisting of sinusoidal positional embeddings, tiny MLPs, and a multi-head mixer.
(b) Architecture of each tiny MLP.}
\label{fig:overview1}
\vspace{-8pt}
\end{figure}

\subsection{Multi-point positional insertion tuning}

MPI tuning inserts a multi-head positional (MHP) encoder, which is a lightweight learnable module that incorporates positional information into latent features.
The MHP encoder produces $N$ output embeddings $\mathcal{P} = \{\bm{p}_{i}\}_{i=1}^{N}$, each of which is added to the latent vanilla features $\bm{h}_{i}(\bm{x})$ as follows:
\begin{align}
\label{eq:hplusp}
\bm{h}^{\prime}_{i}(\bm{x})
=
\bm{h}_{i}(\bm{x}) + \bm{p}_{i},
\end{align}
where $\bm{h}^{\prime}_{i}(\bm{x})$ denotes the adapted latent features.
In the finetuning phase, the adapted features are used instead of the vanilla features, and only the parameters of the inserted MHP encoder are optimized.

\subsection{Architecture}

Figure~\ref{fig:overview1} shows the architecture of the MHP encoder, which consists of the following three components:
1) sinusoidal positional embeddings,
2) tiny MLPs,
and
3) a multi-head mixer.

\noindent \textbf{Sinusoidal positional embeddings~\cite{vaswani2017attention}.}
The input of the MHP encoder is
the sinusoidal positional embeddings
$\bm{e} = (\bm{e}_{1}, \bm{e}_{2}, \cdots, \bm{e}_{L}) \in \mathbb{R}^{D \times L}$, defined by
\begin{align}
e_{l, 2k} = \sin\left(\frac{l}{C^{\frac{2k}{D}}}\right), \quad
e_{l, 2k+1} = \cos\left(\frac{l}{C^{\frac{2k}{D}}}\right),
\end{align}
where
$D$ is the dimension, $l = 0, 1, \cdots, L-1$ is the position index, $k = 0, 2, \cdots, D/2-1$ is the element index,
and $C$ is a constant. We use $D = 64$, $L = 80,000$, and $C = 10,000$ as the default values.

\noindent \textbf{Tiny MLPs.}
The sinusoidal positional embeddings are fed into $M$ tiny MLPs.
As shown in Figure~\hyperref[fig:overview1]{\ref*{fig:overview1}b}, each tiny MLP consists of two blocks of a linear layer, LayerNorm~\cite{ba2016layernorm}, and a Swish-Gated Linear Unit (SwiGLU) activation~\cite{shazeer2020swiglu}.
Both linear layers maintain dimension $D=64$.
This produces
output embeddings $\tilde{\bm{e}}^{(j)} = (\tilde{\bm{e}}_{1}^{(j)}, \tilde{\bm{e}}_{2}^{(j)}, \cdots, \tilde{\bm{e}}_{L}^{(j)}) \in \mathbb{R}^{D \times L}$ for $j = 1, 2, \cdots, M$.

\noindent \textbf{Multi-head mixer.}
Finally, the multi-head mixer produces the embeddings $\mathcal{P} = \{\bm{p}_{i}\}_{i=1}^{N}$ used in Eq.~(\ref{eq:hplusp}) from the embeddings $\mathcal{E} = \{\tilde{\bm{e}}^{(j)}\}_{j=1}^{M}$ obtained from the tiny MLPs.
When $M = N$,
we can straightforwardly map each embedding $\tilde{\bm{e}}^{(j)}$ to its corresponding $\bm{p}_{i}$ using a one-to-one correspondence, such that 
$\bm{p}_{i} = g_{i}(\bm{e}^{(i)})$,
where $g_{i}$ is a simple transformation function such as a linear function.
However, for parameter efficiency, reducing $M$ such that $M < N$ is beneficial.
To this end, the multi-head mixer generates $N$ embeddings through a linear combination of the embeddings in $\mathcal{E}$.
Specifically, it generates $\bm{p}_{i}$ as follows:
\begin{align}
\bm{p}_{i} = g_{i} \left(
\sum_{j=1}^{M} A_{ij} \tilde{\bm{e}}^{(j)} \right),
\end{align}
where $A_{ij} \in \mathbb{R}^{N \times M}$ is a learnable matrix and $g_{i}$ is a linear layer.
Each linear layer is designed to match the shapes of $\bm{p}_{i}$ and $\bm{h}_{i}(\bm{x})$ ti ensure that $\bm{p}_{i}$ can be added to $\bm{h}_{i}(\bm{x})$.

\subsection{Application to Grounding DINO}

\begin{figure}
\centering
\includegraphics[width=\linewidth]{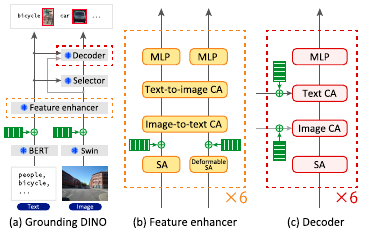}
\vspace{-12pt}
\caption{Application to GDINO. The points to insert embeddings $\bm{p}_{i}$ are highlighted in green. (a) Architecture of GDINO. (b) Architecture of feature enhancer block. (c) Architecture of decoder block.
}
\label{fig:overview2}
\vspace{-8pt}
\end{figure}

This subsection describes the application of MPI tuning to GDINO~\cite{liu2023groundingdino, zhao2024open}, which is the model used in our experiments.
Figure~\ref{fig:overview2} shows the architecture of GDINO, which consists of the following five components:
a BERT (text encoder)~\cite{devlin2018bert},
a Swin transformer (image encoder)~\cite{liu2021swin},
a feature enhancer, a query selector, and a decoder.
Because inserting positional information into all latent features can be redundant due to the complexity of this architecture, we selected $N = 26$ points. These are highlighted in green colors in Figure~\ref{fig:overview2}.

\noindent \textbf{BERT and Swin.}
The first two points correspond to the outputs of the BERT and Swin transformer (Figure~\hyperref[fig:overview2]{\ref*{fig:overview2}a}).
They help learn the positions of the raw input data.

\noindent \textbf{Feature enhancer.}
Each feature enhancer block has two points, one after the self-attention module and the other after the deformable self-attention module, as shown in Figure~\hyperref[fig:overview2]{\ref*{fig:overview2}b}.
This results in twelve points because GDINO has six feature enhancer blocks.

\noindent \textbf{Decoder.}
Each decoder block has two points for the cross-attention module, as shown in Figure~\hyperref[fig:overview2]{\ref*{fig:overview2}c}.
This results in twelve points because GDINO has six decoder blocks.

\subsection{Loss function}

The finetuning loss function is the sum of the localization and counteractive losses as in~\cite{liu2023groundingdino, carion2020detr}.
The detection prompt~\cite{li2021glip, zhang2022glipv2, liu2023groundingdino} that concatenates object category names is used as the text input.
We use the implementation provided with MM-GDINO~\cite{zhao2024open}.

\section{Experiments}

\subsection{Experimental settings}

\noindent \textbf{Datasets.} The SODA-D dataset~\cite{cheng2023soda} was used for finetuning and evaluation. It consists of 24,704 high-quality and high-resolution images of street scenes, along with 277,596 bounding box annotations for small objects across nine object categories.
The official training and test splits were used.
Parameter efficient finetuning experiments were conducted using the MM-Grounding DINO~\cite{zhao2024open} model, which is pretrained on the union of the following four datasets: O365~\cite{Shao2019o365}, GoldG~\cite{kamath2021goldg}, GRIT~\cite{peng2023kosmos} and V3Det~\cite{wang2023v3det}.

\begin{table*}[t]
\centering
\normalsize
\caption{Comparison of small object detection performance on the SODA-D test set.
The proposed method is compared with PEFT methods. For reference, results for zero-shot baseline and full training (Full) methods are also reported. \#Params indicates the number of learnable parameters. \label{tab:main}
}
\setlength{\tabcolsep}{6.7pt}
\begin{tabular}{cl|c|c|ccccccc}
\toprule
\multicolumn{2}{l|}{Method} & Pretrained & \#Params.  & mAP & mAP$_{50}$ & mAP$_{75}$ & mAP$_{eS}$ & mAP$_{rS}$ & mAP$_{gS}$ & mAP$_N$ \\ 
\midrule
\multicolumn{2}{l|}{Zero-shot baseline} & $\checkmark$ & 0 &
14.0 & 31.7 & 10.6 & 3.7 & 10.7 & 18.6 & 27.1\\
\midrule
\multirow{5}{*}{\rotatebox{90}{PEFT}}&
CoOp w/ dec. & $\checkmark$ & 12.00M &
25.8 & 54.7 & 21.2 & 10.5 & 22.1 & 31.6 & 41.8\\
& VPT w/ dec. & $\checkmark$ & 11.98M &
25.4 & 53.7 & 20.9 & 10.0 & 21.6 & 31.2 & 41.5\\
\cmidrule{2-11}
& CoOp w/o dec. & $\checkmark$ & 1.01M &
18.8 & 40.6 & 15.2 & 6.0 & 15.0 & 24.2 & 32.9\\
& VPT w/o dec. & $\checkmark$ & 0.99M &
18.2 & 39.3 & 14.6 & 5.9 & 14.3 & 23.4 & 32.5\\
& Adapter tuning & $\checkmark$ & 0.79M &
22.8 & 49.6 & 18.2 & 8.3 & 19.1 & 28.3 & 38.0\\
& MPI tuning (Ours) & $\checkmark$ & \textbf{0.50M} &
\textbf{25.7} & \textbf{53.7} & \textbf{21.6} & \textbf{9.8} & \textbf{22.1} & \textbf{31.7} & \textbf{41.4}\\
\midrule
\multirow{4}{*}{\rotatebox{90}{Full}}&
Deformable-DETR & & 35.17M & 19.2 & 44.8 & 13.7 & 6.3 & 15.4 & 24.9 & 34.2 \\ 
& Sparse RCNN & & 105.96M & 24.2 & 50.3 & 20.3 & 8.8 & 20.4 & 30.2 & 39.4 \\
& CFINet & & 47.61M & 30.7 & 60.8 & 26.7 & 14.7 & 27.8 & 36.4 & 44.6 \\
& Full fine-tuning & $\checkmark$ & 172.97M & 
32.7 & 64.1 & 29.2 & 15.3 & 28.8 & 39.2 & 50.4\\
\bottomrule
\end{tabular}
\vspace{-5pt}
\end{table*}

\begin{table}[t]
\normalsize
\centering
\begin{minipage}[t]{0.4\linewidth}
\vspace{-5pt}
\caption{Ablation study (validation set). \label{tab:ablation}}
\vspace{-5pt}
\begin{tabular}{l|c}
\toprule
Method  & mAP \\ 
\midrule
MPI tuning & 26.5 \\
w/o input pos. & 26.3 \\
w/o FE pos. & 24.6 \\
w/o dec. pos. & 26.5 \\
\bottomrule
\end{tabular}
\end{minipage}
\hspace{8pt}
\begin{minipage}[t]{0.45\linewidth}
\centering
\vspace{-5pt}
\caption{Hyperparameter study on $M$ (validation set).}
\vspace{-5pt}
\label{tab:hyper}
\setlength{\tabcolsep}{11pt}
\normalsize
\begin{tabular}{c|c|c}
\toprule
$M$ & \#P  & mAP \\
\midrule
24 & 1.01M & \textbf{26.7} \\
12 & 0.50M & 26.5 \\
6 &  0.25M & 26.2 \\
3 &  0.13M & 25.6 \\
\bottomrule
\end{tabular}
\end{minipage}
\end{table}

\noindent \textbf{Evaluation metrics.}
The mean average precision (mAP) computed across multiple intersection over union (IoU) thresholds from 0.50 to 0.95 with an interval of 0.05 was used as a primary evaluation metric, reported along with mAPs at IoU thresholds of 0.50 and 0.75 referred to as mAP$_{50}$ and mAP$_{75}$, respectively.
We also reported mAPs for extremely small objects (mAP$_{eS}$), relatively small objects (mAP$_{rS}$), generally small objects (mAP$_{gS}$), and normal objects (mAP$_{N}$).
To evaluate the parameter efficiency, the number of learnable parameters (\#Params) was reported.

\noindent \textbf{Baselines.}
We selected four baselines: zero-shot detection, CoOp~\cite{zhou2022coop},  VPT~\cite{jia2022vpt} and adapter tuning~\cite{nlp-adapter}.
The zero-shot detection reports the performance before finetuning.
CoOp and VPT are PEFT methods that are based on prompt tuning.
Following~\cite{jia2022vpt}, the head module (the decoder of GDINO) is also finetuned.
The adapter tuning inserts learnable modules to each MLP and self-attention module in the decoder.
Each adapter module consists of two linear layers with a RELU activation in between, followed by LayerNorm.

\noindent \textbf{Implementation details.} 
All the models were trained under the same conditions. Specifically, the AdamW optimizer with a cosine annealing scheduler was used for 12 epochs.
The initial learning rate was set to $10^{-4}$, and the batch size was set to 16.

\subsection{Experimental results}
\noindent \textbf{Main results.}
Table~\ref{tab:main} compares MPI tuning with the conventional PEFT methods.
As shown, it achieved results comparable to CoOp and VPT using a learnable decoder head, while reducing the number of parameters to 0.50 million.
Compared with the zero-shot baseline, the detection performance was significantly improved, highlighting the effectiveness and parameter efficiency of MPI tuning.
Compared with the full training reported as a reference, there is still room for performance improvement. 
Full finetuning of GDINO performed better than CFINet~\cite{yuan2023cfinet}, which is a convolutional neural network designed for small object detection; however, it is parameter inefficient.
To achieve the accuracy of these methods with better parameter efficiency, modules that further enhance the learning efficiency and effectiveness are required in future studies.

\noindent \textbf{Ablation study.}
Table~\ref{tab:ablation} summarizes the results of an ablation study on the incorporation of the positional information.
As shown, incorporating positional information into the feature enhancer was the most effective.
This is because, with GDINO, the fusion of text and image features is the most important process. By inserting a learnable module at this stage, the model can be adapted efficiently to small object detection.

\noindent \textbf{Number of tiny MLPs.}
Table~\ref{tab:hyper} summarizes the results of a hyperparameter study in which the number of tiny MLPs $M$ varies. As shown, larger $M$ yielded better performance.
Setting $M=3$ resulted in a decrease in the performance, but it still significantly outperformed the zero-shot baseline.

\noindent \textbf{Image encoders.}
Table~\ref{tab:backbone} compares the results obtained by three different backbones: Swin-T, Swin-B and Swin-L.
MPI tuning was more parameter-efficient and effective than CoOp without decoder finetuning.

\begin{table}[t]
\normalsize
\centering
\vspace{-8pt}
\caption{Results with different backbones (validation set).}
\vspace{-5pt}
\setlength{\tabcolsep}{3.6pt}
\label{tab:backbone}
\begin{tabular}{cl|c|ccc}
\toprule
\multicolumn{2}{l|}{Method} & \#Params.  & mAP & mAP$_{50}$ & mAP$_{75}$\\ 
\midrule
\multirow{4}{*}{\rotatebox{90}{Swin-T}}&
Zero-shot & -- & 14.3 & 32.1 & 11.0 \\
& CoOp w/ dec. & 12.00M & 26.4 & 56.2 & 21.5 \\
& CoOp w/o dec. & 1.01M & 19.4 & 42.0 & 15.3 \\
& MPI tuning (Ours) & 0.50M & 26.5 & 55.1 & 22.0 \\
\midrule
\multirow{4}{*}{\rotatebox{90}{Swin-B}}&
Zero-shot & -- & 15.1 & 34.2 & 11.5 \\
& CoOp w/ dec. & 12.00M & 27.2 & 57.5 & 22.2 \\
& CoOp w/o dec. & 1.01M & 21.2 & 45.9 & 16.9 \\
& MPI tuning (Ours) & 0.50M & 26.8 & 56.4 & 22.2  \\
\midrule
\multirow{4}{*}{\rotatebox{90}{Swin-L}}&
Zero-shot & -- & 17.6 & 41.3 & 12.9 \\
& CoOp & 12.00M & 30.1 & 61.2 & 25.6 \\
& CoOp w/o dec. & 1.01M & 18.3 & 38.8 & 15.0 \\
& MPI tuning (Ours) & 0.50M & 30.5 & 60.9 & 26.9 \\
\bottomrule
\end{tabular}
\vspace{-12pt}
\end{table}

\begin{figure}
\centering
\vspace{-5pt}
\includegraphics[width=0.99\linewidth]{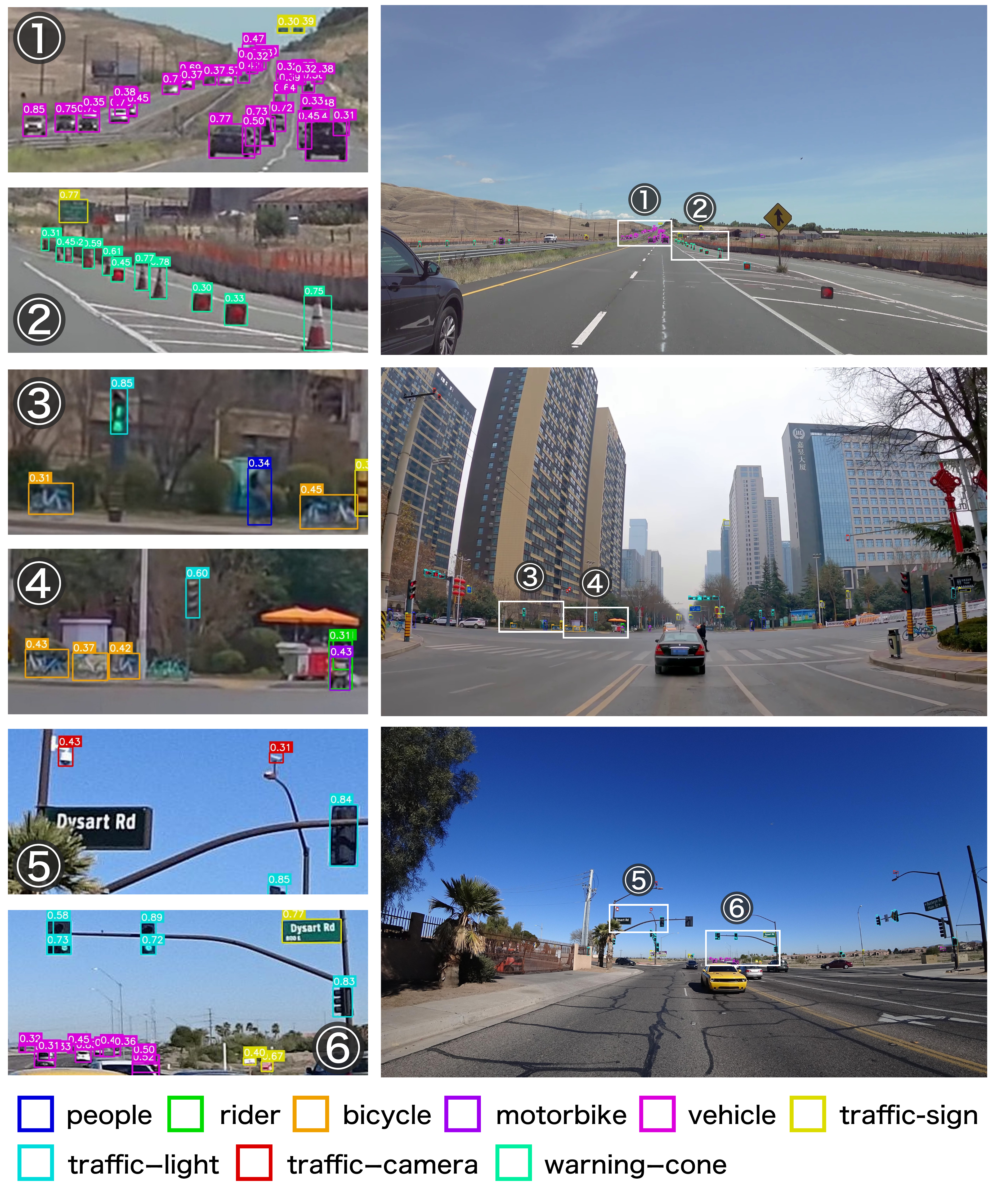}
\vspace{-16pt}
\caption{Qualitative examples}
\label{fig:qualitative}
\vspace{-8pt}
\end{figure}

\noindent \textbf{Qualitative examples.}
Figure~\ref{fig:qualitative} shows qualitative examples.
As can be seen, our method enabled GDINO to detect extremely small objects.

\section{Conclusion}
We proposed MPI tuning, a novel PEFT method for small object detection.
The MHP encoder was introduced to incorporate positional information into the latent features in a frozen pretrained model.
In experiments, MPI tuning was applied to GDINO.
Its effectiveness was demonstrated on the SODA-D dataset in comparison with conventional PEFT methods.

\noindent \textbf{Acknowledgment.} This work was supported by JSPS KAKENHI Grant Numbers 23H00490, 22K12089.
{
\bibliographystyle{IEEEbib}
\bibliography{ref}

\begin{thebibliography}{10}

\bibitem{noh2019bffb}
J. Noh, W. Bae, W. Lee, J. Seo, and G. Kim,
\newblock ``Better to follow, follow to be better: Towards precise supervision of feature super-resolution for small object detection,''
\newblock in {\em IEEE/CVF International Conference on Computer Vision (ICCV)}, 2019, pp. 9725--9734.

\bibitem{bai2018sod-mtgan}
Y. Bai, Y. Zhang, M. Ding, and B. Ghanem,
\newblock ``{SOD-MTGAN}: Small object detection via multi-task generative adversarial network,''
\newblock in {\em European Conference on Computer Vision (ECCV)}, 2018.

\bibitem{yuan2023cfinet}
X. Yuan, G. Cheng, K. Yan, Q. Zeng, and J. Han,
\newblock ``Small object detection via coarse-to-fine proposal generation and imitation learning,''
\newblock in {\em IEEE/CVF International Conference on Computer Vision (ICCV)}, 2023.

\bibitem{kim2021lprmemory}
J.-U. Kim, S. Park, and Y.~M. Ro,
\newblock ``Robust small-scale pedestrian detection with cued recall via memory learning,''
\newblock in {\em IEEE/CVF International Conference on Computer Vision (ICCV)}, 2021, pp. 3030--3039.

\bibitem{wu2020selfmimic}
J. Wu, C. Zhou, Q. Zhang, M. Yang, and J. Yuan,
\newblock ``Self-mimic learning for small-scale pedestrian detection,''
\newblock in {\em ACM International Conference on Multimedia (ACMMM)}, 2020, pp. 2012--2020.

\bibitem{Zhang2023localglobal}
Z. Zhang, P. Gong, H. Sun, P. Wu, and X. Yang,
\newblock ``Dynamic local and global context exploration for small object detection,''
\newblock in {\em IEEE International Conference on Acoustics, Speech and Signal Processing (ICASSP)}, 2023, pp. 1--5.

\bibitem{li2021glip}
L.~H. Li, P. Zhang, H. Zhang, J. Yang, C. Li, Y. Zhong, L. Wang, L. Yuan, L. Zhang, J.-N. Hwang, K.-W. Chang, and J. Gao,
\newblock ``Grounded language-image pre-training,''
\newblock in {\em IEEE/CVF Conference on Computer Vision and Pattern Recognition (CVPR)}, 2022.

\bibitem{zhang2022glipv2}
H. Zhang, P. Zhang, X. Hu, Y.-C. Chen, L.~H. Li, X. Dai, L. Wang, L. Yuan, J.-N. Hwang, and J. Gao,
\newblock ``{GLIPv2}: Unifying localization and vision-language understanding,''
\newblock in {\em Annual Conference on Neural Information Processing Systems (NeurIPS)}, 2022.

\bibitem{liu2023groundingdino}
S. Liu, Z. Zeng, T. Ren, F. Li, H. Zhang, J. Yang, C. Li, J. Yang, H. Su, J. Zhu, et~al.,
\newblock ``Grounding dino: Marrying dino with grounded pre-training for open-set object detection,''
\newblock in {\em European Conference on Computer Vision (ECCV)}, 2024.

\bibitem{zhao2024open}
X. Zhao, Y. Chen, S. Xu, X. Li, X. Wang, Y. Li, and H. Huang,
\newblock ``An open and comprehensive pipeline for unified object grounding and detection,''
\newblock {\em arXiv preprint arXiv:2401.02361}, 2024.

\bibitem{long2024multiway}
Z. Long, G. Killick, R. McCreadie, and G.~A. Camarasa,
\newblock ``Multiway-adapter: Adapting multimodal large language models for scalable image-text retrieval,''
\newblock in {\em IEEE International Conference on Acoustics, Speech and Signal Processing (ICASSP)}, 2024.

\bibitem{zhou2024automatic}
H. Zhou, X. Wan, I. Vulić, and A. Korhonen,
\newblock ``Automatic design of adapter architectures for enhanced parameter-efficient fine-tuning,''
\newblock in {\em IEEE International Conference on Acoustics, Speech and Signal Processing (ICASSP)}, 2024.

\bibitem{zhang2024test}
Y. Zhang and C. Zhang,
\newblock ``Test-time distribution learning adapter for cross-modal visual reasoning,''
\newblock in {\em IEEE International Conference on Acoustics, Speech and Signal Processing (ICASSP)}, 2024.

\bibitem{gao2024adapter}
C. Gao, Q. Xu, P. Qiao, K. Xu, X. Qian, and Y. Dou,
\newblock ``Adapter-based incremental learning for face forgery detection,''
\newblock in {\em IEEE International Conference on Acoustics, Speech and Signal Processing (ICASSP)}, 2024.

\bibitem{nlp-adapter}
N. Houlsby, A. Giurgiu, S. Jastrzebski, B. Morrone, Q. De~Laroussilhe, A. Gesmundo, M. Attariyan, and S. Gelly,
\newblock ``Parameter-efficient transfer learning for nlp,''
\newblock in {\em International Conference on Machine Learning (ICML)}, 2019, pp. 2790--2799.

\bibitem{zhou2022coop}
K. Zhou, J. Yang, C.~C. Loy, and Z. Liu,
\newblock ``Learning to prompt for vision-language models,''
\newblock {\em International Journal of Computer Vision (IJCV)}, 2022.

\bibitem{zhou2022cocoop}
K. Zhou, J. Yang, C.~C. Loy, and Z. Liu,
\newblock ``Conditional prompt learning for vision-language models,''
\newblock in {\em IEEE/CVF Conference on Computer Vision and Pattern Recognition (CVPR)}, 2022, pp. 16816--16825.

\bibitem{jia2022vpt}
M. Jia, L. Tang, B.-C. Chen, C. Cardie, S. Belongie, B. Hariharan, and S.-N. Lim,
\newblock ``Visual prompt tuning,''
\newblock in {\em European Conference on Computer Vision (ECCV)}, 2022.

\bibitem{Lin2024Visual}
P. Lin, Z. Yu, M. Lu, F. Feng, R. Li, and X. Wang,
\newblock ``Visual prompt tuning for weakly supervised phrase grounding,''
\newblock in {\em IEEE International Conference on Acoustics, Speech and Signal Processing (ICASSP)}, 2024.

\bibitem{Xu2024Enhanced}
M. Xu, Z. Guo, Y. Zeng, and D. Xiong,
\newblock ``Enhanced transfer learning with efficient modeling and adaptive fusion of knowledge via prompt tuning,''
\newblock in {\em IEEE International Conference on Acoustics, Speech and Signal Processing (ICASSP)}, 2024.

\bibitem{wang2024cophtc}
F. Cai, Z. Zhang, D. Liu, X. Fang, and J. Tong,
\newblock ``Cophtc: Contrastive learning with prompt tuning for hierarchical text classification,''
\newblock in {\em IEEE International Conference on Acoustics, Speech and Signal Processing (ICASSP)}, 2024.

\bibitem{cheng2023soda}
G. Cheng, X. Yuan, X. Yao, K. Yan, Q. Zeng, X. Xie, and J. Han,
\newblock ``Towards large-scale small object detection: Survey and benchmarks,''
\newblock {\em IEEE Transactions on Pattern Analysis and Machine Intelligence (TPAMI)}, pp. 1--20, 2023.

\bibitem{lin2017retinanet}
T.-Y. Lin, P. Goyal, R. Girshick, K. He, and P. Doll{\'a}r,
\newblock ``Focal loss for dense object detection,''
\newblock in {\em IEEE/CVF International Conference on Computer Vision (ICCV)}, 2017, pp. 2980--2988.

\bibitem{tian2019fcos}
Z. Tian, C. Shen, H. Chen, and T. He,
\newblock ``{FCOS}: Fully convolutional one-stage object detection,''
\newblock in {\em IEEE/CVF International Conference on Computer Vision (ICCV)}, 2019, pp. 9627--9636.

\bibitem{ren2015fasterrcnn}
S. Ren, K. He, R. Girshick, and J. Sun,
\newblock ``{Faster R-CNN}: Towards real-time object detection with region proposal networks,''
\newblock in {\em Annual Conference on Neural Information Processing Systems (NeurIPS)}, 2015.

\bibitem{sun2021sparsercnn}
P. Sun, R. Zhang, Y. Jiang, T. Kong, C. Xu, W. Zhan, M. Tomizuka, L. Li, Z. Yuan, C. Wang, and P. Luo,
\newblock ``{Sparse R-CNN}: End-to-end object detection with learnable proposals,''
\newblock in {\em IEEE/CVF Conference on Computer Vision and Pattern Recognition (CVPR)}, 2021.

\bibitem{carion2020detr}
N. Carion, F. Massa, G. Synnaeve, N. Usunier, A. Kirillov, and S. Zagoruyko,
\newblock ``End-to-end object detection with transformers,''
\newblock in {\em European Conference on Computer Vision (ECCV)}, 2020.

\bibitem{zhu2020deformabledetr}
X. Zhu, W. Su, L. Lu, B. Li, X. Wang, and J. Dai,
\newblock ``{Deformable DETR}: Deformable transformers for end-to-end object detection,''
\newblock in {\em International Conference on Learning Representations (ICLR)}, 2020.

\bibitem{jinyu24icassp}
J. Shi and W. Wu,
\newblock ``Srp-uod: Multi-branch hybrid network framework based on structural re-parameterization for underwater small object detection,''
\newblock in {\em IEEE International Conference on Acoustics, Speech and Signal Processing (ICASSP)}, 2024, pp. 2715--2719.

\bibitem{Li2024SOD}
Y. Li, Y. Wang, Z. Ma, X. Wang, and Y. Tang,
\newblock ``Sod-uav: Small object detection for unmanned aerial vehicle images via improved yolov7,''
\newblock in {\em IEEE International Conference on Acoustics, Speech and Signal Processing (ICASSP)}, 2024.

\bibitem{Zhu2023Small}
J. Zhu, Y. Yang, and Y. Cheng,
\newblock ``Small object detection on the water surface based on radar and camera fusion,''
\newblock in {\em IEEE International Conference on Acoustics, Speech and Signal Processing (ICASSP)}, 2024.

\bibitem{Zhang2023Dynamic}
Z. Zhang, P. Gong, H. Sun, P. Wu, and X. Yang,
\newblock ``Dynamic local and global context exploration for small object detection,''
\newblock in {\em IEEE International Conference on Acoustics, Speech and Signal Processing (ICASSP)}, 2023.

\bibitem{Otake2023Parameter}
S. Otake, R. Kawakami, and N. Inoue,
\newblock ``Parameter efficient transfer learning for various speech processing tasks,''
\newblock in {\em IEEE International Conference on Acoustics, Speech and Signal Processing (ICASSP)}, 2023.

\bibitem{vaswani2017attention}
A. Vaswani, N. Shazeer, N. Parmar, J. Uszkoreit, L. Jones, A.~N. Gomez, L. Kaiser, and I. Polosukhin,
\newblock ``Attention is all you need,''
\newblock in {\em Annual Conference on Neural Information Processing Systems (NeurIPS)}, 2017, pp. 6000--6010.

\bibitem{ba2016layernorm}
J.~L. Ba, J.~R. Kiros, and G.~E. Hinton,
\newblock ``Layer normalization,''
\newblock in {\em NeurIPS Deep Learning Symposium}, 2016.

\bibitem{shazeer2020swiglu}
N. Shazeer,
\newblock ``{GLU} variants improve transformer models,''
\newblock {\em arXiv preprint arXiv:2002.05202}, 2020.

\bibitem{devlin2018bert}
J. Devlin, M.-W. Chang, K. Lee, and K. Toutanova,
\newblock ``Bert: Pre-training of deep bidirectional transformers for language understanding,''
\newblock in {\em Annual Conference of the North American Chapter of the Association for Computational Linguistics (NAACL)}, 2019.

\bibitem{liu2021swin}
Z. Liu, Y. Lin, Y. Cao, H. Hu, Y. Wei, Z. Zhang, S. Lin, and B. Guo,
\newblock ``Swin transformer: Hierarchical vision transformer using shifted windows,''
\newblock in {\em IEEE/CVF International Conference on Computer Vision (ICCV)}, 2021.

\bibitem{Shao2019o365}
S. Shao, Z. Li, T. Zhang, C. Peng, G. Yu, X. Zhang, J. Li, and J. Sun,
\newblock ``Objects365: A large-scale, high-quality dataset for object detection,''
\newblock in {\em IEEE/CVF International Conference on Computer Vision (ICCV)}, 2019.

\bibitem{kamath2021goldg}
A. Kamath et~al.,
\newblock ``Mdetr-modulated detection for end-to-end multi-modal understanding,''
\newblock in {\em IEEE/CVF International Conference on Computer Vision (ICCV)}, 2021, pp. 1780--1790.

\bibitem{peng2023kosmos}
Z. Peng, W. Wang, L. Dong, Y. Hao, S. Huang, S. Ma, and F. Wei,
\newblock ``Kosmos-2: Grounding multimodal large language models to the world,''
\newblock {\em arXiv preprint arXiv:2306.14824}, 2023.

\bibitem{wang2023v3det}
J. Wang, P. Zhang, T. Chu, Y. Cao, Y. Zhou, T. Wu, B. Wang, C. He, and D. Lin,
\newblock ``V3det: Vast vocabulary visual detection dataset,''
\newblock in {\em IEEE/CVF International Conference on Computer Vision (ICCV)}, 2023, pp. 19844--19854.

\end{thebibliography}
}

\end{document}